\documentclass{article}

\usepackage{PRIMEarxiv}

\usepackage[utf8]{inputenc} 
\usepackage[T1]{fontenc}    
\usepackage{hyperref}       
\usepackage{url}            
\usepackage{booktabs}       
\usepackage{amsfonts}       
\usepackage{nicefrac}       
\usepackage{microtype}      
\usepackage{lipsum}
\usepackage{fancyhdr}       
\usepackage{graphicx}       
\usepackage{siunitx}
\usepackage{amsmath}
\usepackage{multirow}
\usepackage{amsfonts}
\usepackage{hyperref}

\title{World Model for AI Autonomous Navigation in Mechanical Thrombectomy
\thanks{\textit{\underline{Citation}}: 
\textbf{Robertshaw, H., Wu, HR., Granados, A., Booth, T.C. (2026). World Model for AI Autonomous Navigation in Mechanical Thrombectomy. In: Gee, J.C., et al. Medical Image Computing and Computer Assisted Intervention – MICCAI 2025. MICCAI 2025. Lecture Notes in Computer Science, vol 15968. Springer, Cham. https://doi.org/10.1007/978-3-032-05114-1\_65}} 
}

\author{
  Harry Robertshaw, Alejandro Granados, Thomas C Booth \\
  School of Biomedical Engineering \& Imaging Sciences \\
  Kings College London \\
  London, UK\\
  \texttt{thomas.booth@kcl.ac.uk} \\
   \And
  Han-Ru Wu \\
  Nuffield Department of Primary Care Health Sciences \\
  University of Oxford \\
  Oxford, UK\\
}

\begin{document}
\maketitle

\begin{abstract}
    Autonomous navigation for mechanical thrombectomy (MT) remains a critical challenge due to the complexity of vascular anatomy and the need for precise, real-time decision-making. Reinforcement learning (RL)-based approaches have demonstrated potential in automating endovascular navigation, but current methods often struggle with generalization across multiple patient vasculatures and long-horizon tasks. We propose a world model for autonomous endovascular navigation using TD-MPC2, a model-based RL algorithm. We trained a single RL agent across multiple endovascular navigation tasks in ten real patient vasculatures, comparing performance against the state-of-the-art Soft Actor-Critic (SAC) method. Results indicate that TD-MPC2 significantly outperforms SAC in multi-task learning, achieving a 65\% mean success rate compared to SAC’s 37\% ($p < 0.001$), with notable improvements in path ratio. TD-MPC2 exhibited increased procedure times, suggesting a trade-off between success rate and execution speed. These findings highlight the potential of world models for improving autonomous endovascular navigation and lay the foundation for future research in generalizable AI-driven robotic interventions.
\end{abstract}


\section{Introduction}

    Stroke is a leading cause of death and disability worldwide, imposing significant strain on healthcare systems and families \cite{Martin2024}. Mechanical thrombectomy (MT) has improved ischemic stroke outcomes, reducing mortality and disability compared to medical therapy alone \cite{Bendszus2023,Nogueira2018}. However, its effectiveness declines with delayed treatment, emphasizing the need for rapid intervention \cite{Nogueira2018,Asdaghi2023}. Despite its efficacy, only 3.1\% of stroke admissions in, for example, the UK receive MT, far below the estimated 15\% eligibility rate, due to limited access to MT-capable centers and long transfer times \cite{SSNAP2023,McMeekin2024,Ngankou2021}. Additionally, operators face radiation exposure, increasing cancer and cataract risks, while protective gear contributes to orthopedic strain \cite{Klein2009,Weyland2020,Madder2017}. Robotic surgical systems offer a potential solution by improving accessibility and reducing operator dependency \cite{Robertshaw2023}. Tele-operated robotic MT could (1) allow specialists in centralized centers to perform procedures remotely, while (2) AI assistance robots may enable less experienced operators (interventional radiologists in peripheral hospitals) to perform MT effectively. Integrating AI into robotic systems could further enhance efficiency and safety, with autonomous surgical robots already demonstrating superior performance over manual techniques in some scenarios \cite{RiveroMoreno2024}.

    Autonomous endovascular navigation research has primarily focused on the aortic arch \cite{Jianu2024,Karstensen2024,Robertshaw2023}, with recent work exploring micro-guidewire and micro-catheter navigation in MT’s second phase \cite{Robertshaw2025}. These studies use multiple test vasculatures, but examine simple tasks with short episodes, limiting clinical applicability. Additionally, autonomous two-device navigation for phase one of MT has been investigated using a larger episode size but was limited to testing on a single vasculature \cite{Robertshaw2024}. These studies use reinforcement learning (RL) algorithms, the majority of which are sensitive to architecture, hyperparameters, and are unable to perform over long time horizons, while often being designed for single-task learning only \cite{Henderson2018,Georgiev2024}. For complex or multiple tasks, this limits RL to computationally expensive models or tasks where tuning is prohibitive \cite{Hafner2023}.

    To move towards realizing the benefits of autonomous MT navigation, it is necessary to develop policies capable of performing long navigation tasks across multiple patients vasculatures that can be adapted efficiently to new robots, tasks, and environments \cite{ONeill2024}. One promising technique may be world models, a learned representation of the environment that simulates its dynamics, enabling a single agent to optimize actions in a virtual setting without relying solely on real-world data \cite{Ha2018}. Using this model, large-scale learning from diverse datasets could create AI navigation systems that can understand, predict, and adapt to real-world complexities \cite{Hu2023}. Recent work has shown that single configurations of RL algorithms (DreamerV3 and TD-MPC2) with no hyperparameter tuning can outperform specialized methods across diverse benchmark tasks \cite{Hafner2023,Hansen2024}. They can also complete farsighted tasks such as collecting diamonds in Minecraft without human data or curricula and capturing expectations of future events during autonomous driving \cite{Hafner2023,Hu2023}. Although the translation of these models to real-world applications is limited, they hold the potential for creating agents capable of performing long navigation tasks on a diverse range of patient anatomies. TD-MPC2 has shown significant improvements upon both Soft Actor-Critic (SAC) (a configuration of which is the current state-of-the-art for autonomous endovascular interventions compared to benchmarks \cite{Karstensen2024,Moosa2025}), and DreamerV3 when examining multi-task environments with continuous action spaces. It is also able to utilize multiple data types, such as human demonstrator, RL collected, and its own online interaction data (from a single task or across multiple tasks).

    The aim of this study was to propose a framework for a world model for autonomous endovascular interventions, and more specifically, MT. The primary objective was to demonstrate that a singular RL agent could be used to perform multiple endovascular navigation tasks across multiple patient vasculatures. Our contributions are as follows: 1) we implemented a world model capable of performing multiple endovascular navigation tasks across multiple patient vasculatures, for the first time, 2) we proposed a framework for endovascular intervention multi-task learning, 3) we compared our results to the current state-of-the-art RL algorithms for autonomous endovascular interventions, demonstrating performance across the largest dataset examined for autonomous MT.

\section{Methods}

    \subsection{Navigation tasks}

        MT typically involves navigating a guide catheter from the femoral or radial artery to the internal carotid artery (ICA). An `access catheter' is usually placed within the guide catheter and taken ahead of the guide catheter tip during navigation. Once the access catheter is within the carotid artery, the guide catheter is advanced to make a stable platform. At this point, the guidewire and access catheter are retracted, and a micro-guidewire within a micro-catheter is passed through the stable guide catheter and navigated to the target thrombus site. Frequently, the final step is to remove the micro-guidewire and exchange it for a stent retriever to remove the thrombus, restoring blood flow to the brain.

\begin{figure}[hbt!]
            \centering
            \includegraphics[width=0.50\linewidth]{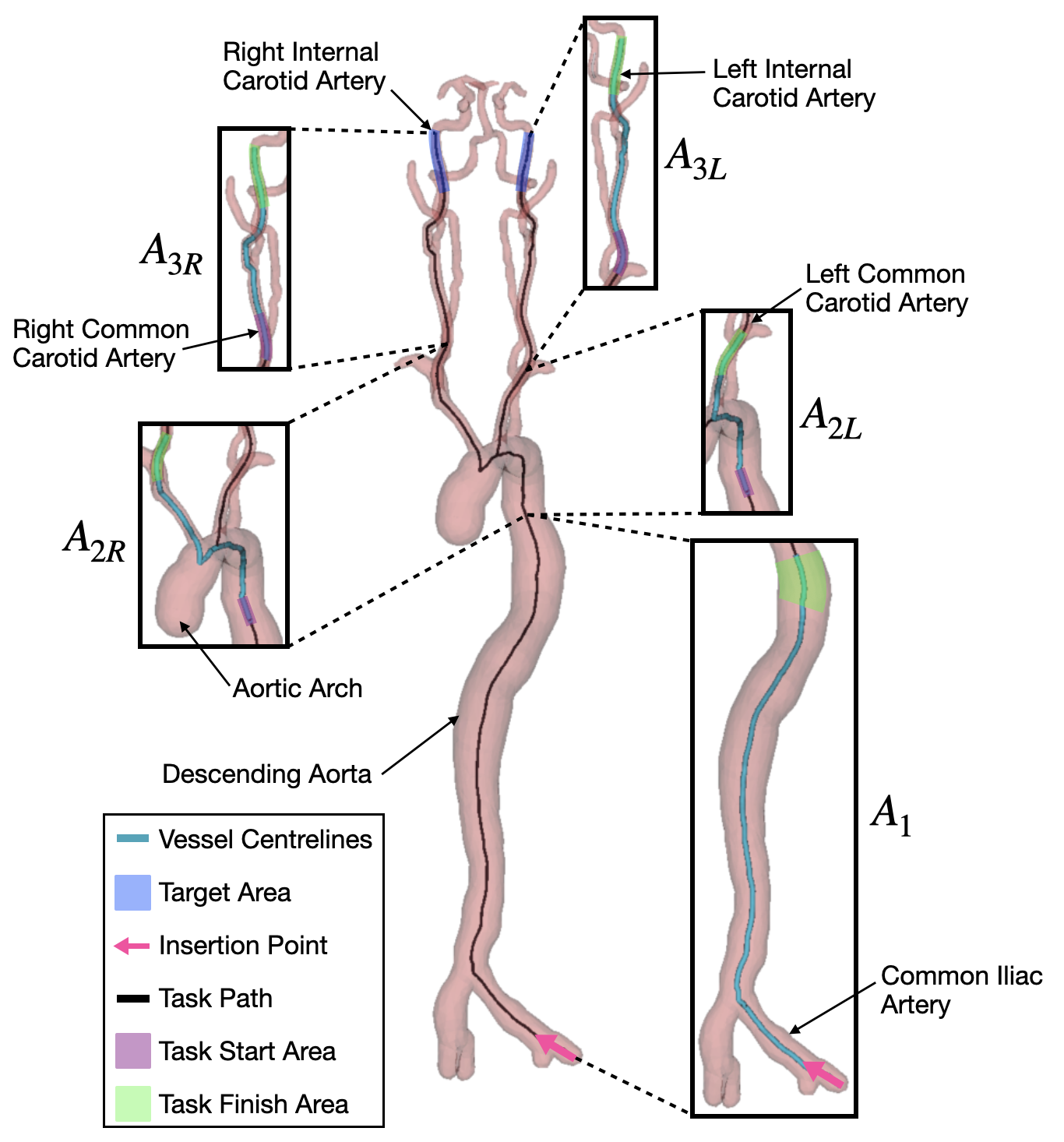}
            \caption{Full MT vasculature with each navigation task labeled. $A_1$: Common iliac artery to top of descending aorta, $A_{2L}$: top of descending aorta to left common carotid artery (CCA), $A_{2R}$: top of descending aorta to right CCA, $A_{3L}$: left CCA to left ICA, $A_{3R}$: right CCA to right ICA.}
            \label{fig:task_diagram}
        \end{figure}

        The first phase of MT (navigation of guidewire and guide catheter from the femoral to the ICA) has been split into five tasks for multi-task RL training. These are based on previously defined phases of MT \cite{Crossley2019} and are described in Figure~\ref{fig:task_diagram}. A target was randomly sampled from the set of points within the target vessel, and guide catheter and guidewire navigated to it from a randomly sampled point in the starting vessel.

        The device's distal position was described by three points equally spaced 2\,\unit{\milli\meter} apart along the guidewire, denoted as $(x_g', y_g')_{i=1,2,3}$, with $(x_g', y_g')_{1}$ representing the instrument tip. The target location was specified by the coordinates $(x_t', y_t')$. Observations comprised current and previous device positions, target location, and the previous action taken. 

    \subsection{Data collection}

        The \textit{in silico} environment for the navigation task utilized the stEVE framework (Available at: \url{https://github.com/lkarstensen/stEVE} \cite{Karstensen2024}). The BeamAdapter plugin for Simulation Open Framework Architecture (SOFA, v23.06) was used to model the Neuron MAX 0.088" guide catheter (Penumbra, California, USA) (160 vertices with Young's modulus of 47\,\unit{\mega\pascal}) and 0.035" guidewire (Terumo, Tokyo, Japan) (120 vertices with Young's modulus of 43\,\unit{\mega\pascal}) used for navigation \cite{Wei2012,Faure2012}. The simulation assumed rigid vessel walls (with an empty lumen). The simulation's real-world device behavior was determined by utilizing a tensile testing machine to measure the tensile strength of the devices, which facilitated the calculation of their stiffness. Friction between the wall and guidewire was iteratively tuned to mimic guidewire behavior in a test-bench set-up. Such methodology previously allowed the translation of autonomous endovascular navigation agents from \textit{in silico} to \textit{in vitro} aortic arch models and from \textit{in silico} to \textit{ex vivo} porcine liver vasculatures \cite{Karstensen2024,Karstensen2022}. 

        A computed tomography angiography (CTA) scan which encompassed the aortic arch and extended to the cerebral vessels, and a CTA scan encompassing the abdominal and thoracic regions, including the femoral arteries, descending aorta, and the aortic arch (obtained with UK Research Ethics Committee 24/LO/0057), were processed into surface meshes using segmentation tools in 3D Slicer (v5.8.0). The centerlines with radii for all scans were then extracted. The centerlines of the abdominal and thoracic regions were then scaled appropriately so that the radius at the last point of the descending aorta would match the first point on the aortic arch centerline. These two sets of centerlines were then joined, and the radius at each centreline point was used to generate a surface mesh, which was loaded into SOFA. This was repeated for ten aortic arch and cerebral vessel CTAs, with the same CTA of the abdominal and thoracic regions being scaled to fit each one. The mean right and left-hand-side vessel tortuosity were $1.19\pm0.07$~\unit{\milli\meter} and $1.14\pm0.05$~\unit{\milli\meter}, respectively. Type-I aortic arches were found in 80\% (8/10) of vasculatures, while the remaining were Type-II. Additional augmentation of the entire mesh was applied during training via random scaling (0.7 to 1.3 for height and width) to enhance generalization.  

        Input parameters, including device rotation and translation speed, were applied at the proximal device end. Rotation and translation speed was constrained to a maximum of 180\,\unit{\degree\per\second} and 40\,\unit{\mm\per\second}, respectively. Similar to a clinical scenario with fluoroscopy, feedback during the navigation was given as two-dimensional $(x',y')$ tracking coordinates of three points along each device's tip; no visual information showing the geometry of the patient vasculature was given.

    \subsection{Controller architecture}

        Two types of RL agents were trained in this study. One type used an implementation of the model-free RL algorithm, SAC from the stEVE framework \cite{Karstensen2024}, whereby the critic learns the value and the actor optimizes the critic directly to maximize cumulative rewards \cite{Georgiev2024}. This is useful for continuous action spaces, leveraging experience replay for data efficiency, but requires tuning and struggles under high-dimensional inputs \cite{Hafner2023}. The architecture includes a Long Short-Term Memory (LSTM) layer for learning trajectory-dependent state representations and feedforward layers for controlling the devices. The controller takes observations as input, and a Gaussian policy network outputs mean ($\mu$) and standard deviation ($\sigma$) for expected actions, representing the micro-catheter and micro-guidewire rotations and translations. During training, actions are sampled from the $\sigma$, but for evaluation, $\mu$ is used directly for deterministic behavior.

        The other type of agent was trained using TD-MPC2, a model-based RL algorithm designed for sample-efficient learning and effective planning in continuous action spaces (Available at: \url{https://github.com/nicklashansen/tdmpc2} \cite{Hansen2024}). TD-MPC2 combines temporal difference learning with model-predictive control, leveraging a learned dynamics model to simulate environment transitions and plan action sequences over a short horizon. This approach enables the agent to optimize actions based on both predicted rewards and task constraints while using limited real-world interaction. By incorporating a latent dynamics model and a cross-entropy planning method, TD-MPC2 is particularly well-suited for complex tasks with longer episode lengths. This study used the base TD-MPC2 configuration but increased the length of the replay buffer from $1\times10^6$ to $1\times10^7$. Additionally, we implemented an LSTM layer as an observation embedder, allowing storage of the environment state at each step so that the vessel structure could be estimated by the path of the device tip.

        The dense reward function used during training in both models is shown in Equation~\ref{eq:R}. \textit{Pathlength} is defined as the distance between the guidewire tip and the target along the centerlines of the arteries, with $\Delta\text{pathlength}$ representing the change in pathlength at time $t$ from the previous step at time $t=-1$.

        \begin{equation}
            R = -0.00015 - 0.001\cdot\Delta\text{pathlength}+\begin{cases}1 & \text{if target reached} \\0 & \text{else} \end{cases}
            \label{eq:R}
        \end{equation}

    \subsection{Training and evaluation}

        Initially, five agents were trained using SAC (one for each defined phase in Figure~\ref{fig:task_diagram}). Each of these agents was then evaluated for 250 episodes while the trajectory data was recorded. This data was used to fill the replay buffer of the SAC world model and the TD-MPC2 world model before initiating training. All models were trained for $1 \times 10^7$ exploration steps, with each navigation task (or episode) considered complete when the target was reached within a threshold equal to the entire cross-section of the vessel. An episode termination of 200 steps (approximately 27\,\unit{\second}) was set for computational efficiency. Training was performed on an NVIDIA DGX A100 node (Santa Clara, California, USA).

        We train and test on 10 patient-specific vascular anatomies, with evaluations conducted every $2.5 \times 10^5$~exploration steps for 250~episodes, recording the success rate, procedure time, and path ratio. Comparative statistical analyses were conducted using two-tailed paired Student's t-tests with a predetermined significance threshold set at $p = 0.05$.

\section{Results}

    \subsection{Single task training}

        Results from single-task training for each of the five MT phases can be seen in Table~\ref{tab:single}. Success rates of 100\% were recorded for phases $A_1$ and $A_{2R}$, while high success rates and path ratios were exhibited in phase $A_{2L}$ (SR: $82\%$, PR: $90\%$), $A_{2R}$ (SR: $90\%$, PR: $90\%$), and $A_{3L}$ (SR: $92\%$, PR: $93\%$). This data was used to fill replay buffers before training for the world models in Section~\ref{world_model}.

        \begin{table}[hbt!]
            \centering
            \scriptsize
            \caption{Single task training using SAC. \textit{Success Rate}: percentage of evaluation episodes where the target is reached. \textit{Path Ratio}: remaining distance to the target point in unsuccessful episodes, calculated by dividing the remaining distance by the initial distance. \textit{Procedure Time}: time from the start of navigation to the target location for successful episodes. \textit{Exploration Steps}: number of training steps taken to reach the point at which the results are provided. The reported values are mean $\pm$ standard deviation (standard deviation values may exceed logical bounds (0–100\%) due to the statistical calculation).}
            \label{tab:single}
            \begin{tabular}{c|c|c|c|c}
            \textbf{Task} & \textbf{\begin{tabular}[c]{@{}l@{}}Success Rate (\%)\end{tabular}} & \textbf{\begin{tabular}[c]{@{}l@{}}Procedure \\ Time (s)\end{tabular}} & \textbf{\begin{tabular}[c]{@{}l@{}}Path Ratio (\%)\end{tabular}} & \textbf{\begin{tabular}[c]{@{}l@{}}Exploration \\ Steps\end{tabular}} \\ \hline
            $A_1$ & $100 \pm 0$ & $9.8 \pm 1.5$ & \begin{tabular}[c]{@{}l@{}}$100 \pm 0$\end{tabular} & $0.25\times10^6$ \\ \hline
            $A_{2L}$ & $82 \pm 39$ & $12.3 \pm 9.9$ & \begin{tabular}[c]{@{}l@{}} $90 \pm 18$\end{tabular} & $8.5\times10^6$ \\ \hline
            $A_{2R}$ & $100 \pm 0$ & $10.7 \pm 7.4$ & \begin{tabular}[c]{@{}l@{}}$100 \pm 0$\end{tabular} & $9.5\times10^6$ \\ \hline
            $A_{3L}$ & $90 \pm 30$ & $8.7 \pm 6.1$ & \begin{tabular}[c]{@{}l@{}}$90 \pm 26$\end{tabular} & $2.5\times10^6$ \\ \hline
            $A_{3R}$ & $92 \pm 27$ & $10.9 \pm 6.9$ & \begin{tabular}[c]{@{}l@{}}$93 \pm 17$\end{tabular} & $7.0\times10^6$ \\ 
            \end{tabular} 
        \end{table}

    \subsection{World model}\label{world_model}

        Results from training on multiple vasculatures and navigation tasks showed a significant increase in mean success rate across all tasks for TD-MPC2, 65\% compared to 37\% for the current state-of-the-art (SAC) ($p< 0.001$). These results can be seen in Table~\ref{tab:world_task}. Furthermore, significant increases in success rate and path ratio were seen across tasks when moving from SAC to TD-MPC2: $A_{2R}$ (SR: $8\%$ to $66\%$ [$p< 0.001$], PR: $29\%$ to $77\%$ [$p< 0.001$]), $A_{3L}$ (SR: $16\%$ to $50\%$ [$p< 0.001$], PR: $26\%$ to $6\%$ [$p< 0.001$]), and $A_{3R}$ (SR: $56\%$ to $90\%$ [$p< 0.001$], PR: $63\%$ to $90\%$ [$p= 0.001$]). Additionally, an increase in procedure time was observed across all tasks when moving from SAC to TD-MPC2, with significance in $A_{1}$ (PT: 8.9\,\si{\second} to 16.7\,\si{\second} [$p< 0.001$]), $A_{3L}$ (PT: 7.2\,\si{\second} to 13.7\,\si{\second} [$p= 0.021$]), and $A_{3L}$ (PT: 5.6\,\si{\second} to 12.9\,\si{\second} [$p< 0.001$]). Exploration steps recorded for SAC and TD-MPC2 were $4.5\times10^6$ (reached after 75~hours) and $0.5\times10^6$ (reached after 25~hours), respectively. 

        \begin{table}[hbt!]
            \centering
            \scriptsize
            \caption{World model results per navigation task, with mean results across all tasks and vasculatures.}
            \label{tab:world_task}
            \begin{tabular}{c|ccc|ccc}
            \multicolumn{1}{c|}{\multirow{2}{*}{\textbf{Task}}} & \multicolumn{3}{c|}{\textbf{SAC}} & \multicolumn{3}{c}{\textbf{TD-MPC2}} \\ \cline{2-7} 
            \multicolumn{1}{c|}{} & \multicolumn{1}{l|}{\textbf{\begin{tabular}[c]{@{}l@{}}Success \\ Rate (\%)\end{tabular}}} & \multicolumn{1}{l|}{\textbf{\begin{tabular}[c]{@{}l@{}}Procedure \\ Time (s)\end{tabular}}} & \textbf{\begin{tabular}[c]{@{}l@{}}Path \\ Ratio (\%)\end{tabular}} & \multicolumn{1}{l|}{\textbf{\begin{tabular}[c]{@{}l@{}}Success \\ Rate (\%)\end{tabular}}} & \multicolumn{1}{l|}{\textbf{\begin{tabular}[c]{@{}l@{}}Procedure \\ Time (s)\end{tabular}}} & \textbf{\begin{tabular}[c]{@{}l@{}}Path \\ Ratio (\%)\end{tabular}} \\ \hline
            $A_1$ & \multicolumn{1}{l|}{\textbf{96 $\pm$ 20}} & \multicolumn{1}{l|}{\textbf{8.9 $\pm$ 1.6}} & $89 \pm 16$ & \multicolumn{1}{l|}{\textbf{96 $\pm$ 20}} & \multicolumn{1}{l|}{$16.7 \pm 4.1$} & \textbf{90 $\pm$ 3} \\ \hline
            $A_{2L}$ & \multicolumn{1}{l|}{$10 \pm 30$} & \multicolumn{1}{l|}{\textbf{14.5 $\pm$ 7.1}} & $27 \pm 30$ & \multicolumn{1}{l|}{\textbf{22 $\pm$ 41}} & \multicolumn{1}{l|}{$17.0 \pm 6.1$} & \textbf{45 $\pm$ 33} \\ \hline
            $A_{2R}$ & \multicolumn{1}{l|}{$8 \pm 27$} & \multicolumn{1}{l|}{\textbf{12.8 $\pm$ 9.9}} & $29 \pm 27$ & \multicolumn{1}{l|}{\textbf{66 $\pm$ 47}} & \multicolumn{1}{l|}{$16.5 \pm 6.2$} & \textbf{77 $\pm$ 32} \\ \hline
            $A_{3L}$ & \multicolumn{1}{l|}{$16 \pm 37$} & \multicolumn{1}{l|}{\textbf{7.2 $\pm$ 7.0}} & $26 \pm 34$ & \multicolumn{1}{l|}{\textbf{50 $\pm$ 50}} & \multicolumn{1}{l|}{$13.7 \pm 6.3$} & \textbf{66 $\pm$ 38} \\ \hline
            $A_{3R}$ & \multicolumn{1}{l|}{$56 \pm 50$} & \multicolumn{1}{l|}{\textbf{5.6 $\pm$ 4.5}} & $63 \pm 43$ & \multicolumn{1}{l|}{\textbf{90 $\pm$ 30}} & \multicolumn{1}{l|}{$12.9 \pm 5.8$} & \textbf{90 $\pm$ 21} \\ \hline\hline
            Mean & \multicolumn{1}{l|}{37 $\pm$ 48} & \multicolumn{1}{l|}{\textbf{8.2 $\pm$ 4.6}} & $47 \pm 40$ & \multicolumn{1}{l|}{\textbf{65 $\pm$ 48}} & \multicolumn{1}{l|}{$15.2 \pm 5.8$} & \textbf{73 $\pm$ 33} \\
            \end{tabular}
        \end{table}

\section{Discussion}

    The results of this study demonstrate notable progress in the development of RL models for autonomous MT, addressing multiple navigation tasks across ten real patient vasculatures in a clinically-based simulation using industry-standard devices. This represents the largest dataset examined for autonomous MT, surpassing previous studies that focused on limited anatomical variations and smaller-scale evaluations \cite{Robertshaw2024,Karstensen2024}. The results from single-task training indicate that SAC performs well in short tasks, achieving success rates of 100\% in phases $A_1$ and $A_{2R}$. Other tasks, such as $A_{2L}$, $A_{3L}$, and $A_{3R}$, exhibited moderate to high success rates, with corresponding path ratios. These results allowed suitable data to be used for inputs to the world model, and illustrate SAC's ability to effectively learn task-specific policies with high levels of accuracy and path efficiency when trained on shorter tasks. The reduced performance on left-sided navigation tasks ($A_{2L}$ and $A_{3L}$) highlights potential challenges related to increased vasculature complexity. This variation requires further exploration to identify anatomical features that influence performance but reflects clinical scenarios where the left CCA is typically harder to catheterize and may need specialist access catheters.

    The world model results demonstrate a clear advantage of TD-MPC2 over SAC. A statistically significant increase in mean success rate was observed, with TD-MPC2 achieving 65\% compared to 37\% for SAC ($p < 0.001$). Notably, TD-MPC2 significantly outperformed SAC in $A_{2R}$, $A_{3L}$, and $A_{3R}$, while recording the same or non-significant increases in success rate for $A_{1}$ and $A_{2L}$. This highlights the advantages of TD-MPC2’s model-based framework, leveraging a learned dynamics model and horizon-based planning to optimize actions in diverse and complex environments. SAC’s model-free approach may struggle to generalize effectively to anatomical variations and multiple objectives. However, in this context and other work, TD-MPC2 failed to solve all tasks it was trained on and exhibits limited scalability due to online planning \cite{Georgiev2024}. The findings suggest that TD-MPC2 is better suited for scenarios requiring multi-task generalization and adaptability to unseen environments. In TD-MPC2, the lower success rates seen in $A_{2L}$ (22\%) and $A_{3L}$ (50\%) compared to other tasks may be attributed to the lower performance of the corresponding single-task models. Despite gains in success rate and path ratio, TD-MPC2 resulted in increased procedure time across all tasks. The longer procedure times suggest that TD-MPC2 may conceivably prioritize exploration and more deliberate navigation strategies, leading to improved performance but at the cost of increased procedure time. Future work could explore optimizing the trade-off between efficiency and exploration in navigation.
    
    These results underscore the importance of scalable and generalizable RL approaches for realistic clinical applications of autonomous MT. While the success rate of TD-MPC2 (65\%) highlights the need for further optimization, it represents a significant step toward bridging the gap between experimental performance and clinical feasibility. The comparison of single and multi-task training reveals critical trade-offs between task specialization and generalization. SAC demonstrated high success rates in single-task training, but limited applicability to real-world clinical scenarios, where agents must navigate diverse vasculatures. Conversely, TD-MPC2’s superior performance in multi-task settings indicates its potential as a robust framework for autonomous MT navigation, though its performance remains suboptimal for clinical deployment. For world models, performance relies on pre-existing data to train the world model, which might not always be feasible, especially in novel environments such as this one. Here we propose a balance between number of tasks and a single agent to execute them.

    Future work should focus on improving TD-MPC2’s training pipeline to improve generalization to unseen diverse vasculatures. Current world models have demonstrated scalability to 80-150 tasks \cite{Hafner2023,Georgiev2024,Hansen2024}, underscoring their potential for future work as it looks to incorporate more diverse datasets, more detailed simulation environments that consider contact forces \cite{Robertshaw2025}, while transitioning to \textit{in vitro} testbeds. Although the vessels navigated in $A_{1}$ and $A_{2}$ are almost entirely rigid in practice, vessel deformation and motion for $A_{3}$ may enhance \textit{in vitro} translation. Using differentiable physics simulators could further enable efficient policy learning using First-order Gradients, while work is needed to provide gradient stability over long horizons. Future work may also compare image-based RL, such as DreamerV3~\cite{Hafner2023}, to the current approach, to evaluate if human-like visual input provides benefit. Additionally, balancing task-specific performance with generalization should remain a priority for advancing the clinical utility of AI-based autonomous endovascular navigation, paving the way for safer and more efficient outcomes in real-world applications.

\section{Conclusion}

    This study demonstrates the feasibility of a world model-based RL approach for autonomous MT, addressing key challenges in generalization and multi-task learning. The proposed world model leverages existing open-source repositories to maximize reproducibility \cite{Karstensen2024,Hansen2024}. Through evaluations across multiple patient vasculatures, TD-MPC2 achieved superior generalization compared to SAC, with a significant increase in success rate and path ratio. Future research should focus on refining world models for more efficient planning, integrating differentiable physics simulators to enhance gradient stability, and expanding datasets for better generalization across diverse vasculatures. The findings presented here provide a critical step toward the development of AI-driven autonomous robotic endovascular interventions with improved safety and precision.

\section*{Acknowledgments}
    Partial financial support was received from the WELLCOME TRUST (Grant Agreement No 203148/A/16/Z), the Engineering and Physical Sciences Research Council Doctoral Training Partnership (Grant Agreement No EP/R513064/1), and the MRC IAA 2021 Kings College London (MR/X502923/1). For the purpose of Open Access, the Author has applied a CC BY public copyright license to any Author Accepted Manuscript version arising from this submission.

\bibliographystyle{unsrt}  
\bibliography{references}

\end{document}